\documentclass[pdflatex,sn-mathphys-num]{sn-jnl}

\usepackage{graphicx}%
\usepackage{multirow}%
\usepackage{amsmath,amssymb,amsfonts}%
\usepackage{amsthm}%
\usepackage{mathrsfs}%
\usepackage[title]{appendix}%
\usepackage{xcolor}%
\usepackage{textcomp}%
\usepackage{manyfoot}%
\usepackage{booktabs}%
\usepackage{mhchem}
\usepackage{algorithm}%
\usepackage{algorithmicx}%
\usepackage{algpseudocode}%
\usepackage{listings}%
\usepackage{makecell}
\usepackage{booktabs} %
\usepackage{tabularx} %
\usepackage{array}
\usepackage{bm}

\usepackage[most]{tcolorbox}
\usepackage{listings}
\usepackage{xcolor}

\lstdefinestyle{promptstyle}{
    basicstyle=\ttfamily\footnotesize,
    breaklines=true,
    breakatwhitespace=false,
    columns=fullflexible,
    keepspaces=true,
    showstringspaces=false,
    frame=none
}

\newtcblisting{promptbox}[2][]{
    enhanced,
    breakable,
    colback=gray!3,
    colframe=gray!60,
    boxrule=0.5pt,
    arc=1mm,
    left=1.5mm,
    right=1.5mm,
    top=1mm,
    bottom=1mm,
    title={#2},
    fonttitle=\bfseries,
    listing only,
    listing options={style=promptstyle},
    #1
}

\usepackage{color}
\usepackage{xcolor}
\usepackage{caption}
\usepackage{subfigure}
\usepackage[table]{xcolor}
\usepackage{tcolorbox}
\definecolor{lightblue}{HTML}{F0F8FF}
\definecolor{top1}{rgb}{0.68, 0.85, 0.9}
\definecolor{top2}{rgb}{0.78, 0.93, 0.96}
\definecolor{top3}{rgb}{0.88, 0.97, 0.99}

\theoremstyle{thmstyleone}

\newcommand{\rre}[1]{{\color{black}#1}}

\theoremstyle{thmstyletwo}

\theoremstyle{thmstylethree}

\begin{document}

\title[Article Title]{\rre{Large language model agents accelerate inverse design of metal–organic frameworks for gas separation}}

\author[1]{\fnm{\rre{Zhaolin}} \sur{Hu}}
\author[1,3]{\fnm{\rre{Hehe}} \sur{\rre{Fan}}}
\author[1]{\fnm{\rre{Wangyihan}} \sur{Guo}}
\author[2]{\fnm{\rre{Meng}} \sur{Xu}}
\author[2]{\fnm{\rre{Chenhao}} \sur{Rao}}
\author[2]{\fnm{\rre{Qiwei}} \sur{Yang}}
\author[1,3]{\fnm{\rre{Yi}} \sur{Yang}}

\affil[1]{\orgdiv{College of Artificial Intelligence}, \orgname{Zhejiang University }, \orgaddress{\city{Hangzhou}, \country{China}}}
\affil[2]{\orgdiv{College of Chemical and Biological Engineering}, \orgname{Zhejiang University } \orgaddress{\city{Hangzhou}, \country{China}}}
\affil[3]{\orgdiv{Institute of Fundamental and Transdisciplinary Research}, \orgname{Zhejiang University } 
\orgaddress{\city{Hangzhou}, \country{China}}}

\abstract{
Metal-organic frameworks (MOFs) offer a highly modular platform for adsorptive gas separation, yet their vast reticular design space makes inverse design difficult under simultaneous constraints of chemical validity, separation performance, and structural diversity. Here, we present LEMO Agent, a large-language-model agent framework for closed-loop inverse design of gas-separation MOFs in MOFid space. LEMO Agent couples language-based candidate generation with MOFid standardization, explicit validity checking, Transformer-based property prediction, structured design memory, and multi-island exploration. Through iterative generate--validate--evaluate--remember cycles, the agent uses feedback from both successful and failed candidates to guide chemically constrained search across linker, metal, and topology choices. We evaluate LEMO Agent on CH$_4$/N$_2$ and CO$_2$/N$_2$ separation tasks. Compared with representative generative, optimization, and agentic baselines, LEMO Agent enriches high-performing candidates, improves predicted separation performance, and maintains broad chemical and topological diversity. Selected candidates are further reconstructed, evaluated by GCMC simulations, and passed through an experimental down-selection workflow based on chemical feasibility and ligand purchasability, leading to initial wet-lab synthesis and SEM characterization. These results demonstrate that large language model agents can serve as interpretable and scalable design engines for accelerating MOF discovery beyond conventional fixed-library screening.
}


\maketitle

\section{Introduction}\label{sec: introduction}

Metal–organic frameworks (MOFs) are crystalline porous solids assembled from metal ions or clusters and multitopic organic linkers through coordination-driven self-assembly and reticular chemistry~\cite{furukawa2013chemistry,zhou2012introduction,eddaoudi2001modular,kitagawa2004functional}. The modular combination of metal nodes and organic linkers enables systematic control over pore size, pore shape, framework topology, surface functionality, and local adsorption environments, giving rise to one of the most chemically diverse families of porous materials~\cite{furukawa2013chemistry,eddaoudi2001modular,kitagawa2004functional}. Owing to their high internal surface areas, ordered pore networks, and tunable host–guest interactions, MOFs have been widely explored for gas storage, adsorption, separation, catalysis, sensing, and related energy and environmental applications ~\cite{furukawa2013chemistry,zhou2012introduction,li2012metal,li2014porous,kreno2012metal,corma2010engineering}. Among these applications, adsorptive gas separation is particularly sensitive to the molecular-level matching between framework pores and gas molecules. In this context, separation performance is not governed by a single descriptor, but by the coupled effects of pore aperture, pore geometry, framework polarity, open metal sites, linker functional groups, framework flexibility, and the spatial distribution of adsorption sites~\cite{li2012metal,li2014porous,kreno2012metal,corma2010engineering,li2009selective,zhao2018metal,lin2019exploration,lin2020microporous}. These characteristics make MOFs an attractive platform for separation-oriented materials design, while also creating a complex chemical and structural search space in which high-performing candidates are difficult to identify by intuition or trial-and-error alone.

The same structural tunability that makes MOFs attractive for separation also creates a formidable combinatorial design problem. The possible combinations of metal nodes, organic linkers, functional groups, network topologies, and interpenetration patterns far exceed the number of MOFs that have been experimentally synthesized or computationally evaluated. Over the past decade, high-throughput computational screening has substantially accelerated MOF discovery by constructing and evaluating large experimental and hypothetical databases, including computation-ready experimental MOF collections, hypothetical MOF libraries, topology-guided MOF databases, and descriptor-rich resources such as ARC-MOF~\cite{wilmer2012large,chung2014computation,burner2023arc,anderson2019increasing,colon2017topologically,chung2019advances}. These efforts have enabled systematic ranking of large candidate pools and have provided valuable structure–property insights for adsorption and separation applications. However, most screening workflows are inherently constrained by the structures already present in a database or by the building blocks, topologies, and assembly rules used during enumeration. As a result, they are powerful for identifying promising materials from a predefined candidate set, but they do not directly address the inverse-design question of how to actively generate new, valid, chemically plausible, and task-optimized MOFs beyond the initial search pool. This limitation motivates the development of adaptive design strategies that can propose new candidates and improve them iteratively according to target separation objectives.

Machine learning has increasingly shifted MOF discovery from passive screening toward predictive and generative design. On the prediction side, crystal graph neural networks and MOF specific Transformer models have enabled rapid estimation of adsorption, diffusion, electronic, and other framework properties from crystal graphs, multimodal structural representations, or compact string representations such as MOFid~\cite{xie2018crystal,kang2023multi,bucior2019identification,cao2023moformer}. On the generation side, reinforcement learning, autoregressive MOFid sequence generation, diffusion models, and flow matching approaches have recently been explored for de novo MOF design, covering property-guided sequence generation, linker generation, pore shape conditioned generation, coarse-grained framework generation, and building block-aware three-dimensional structure generation~\cite{park2024inverse,park2024generative,fu2024mofdiff,park2025multi,badrinarayanan2025mofgpt,kim2025mofflow}. These studies demonstrate the potential of deep generative models to explore reticular chemical space beyond predefined screening libraries. More recently, large language models~\cite{brown2020language,achiam2023gpt,singh2025openai} have shown increasing promise in scientific design tasks because they can use chemical knowledge, literature-derived heuristics, natural language rules, and feedback from previous attempts within a unified reasoning context~\cite{m2024augmenting,boiko2023autonomous,kang2024chatmof,lee2026simmof,ramos2025review}. This capability is particularly useful for MOF design, where candidate generation requires coordinated choices of organic linkers, metal nodes, topology labels, functional groups, and target properties. Compared with conventional generative models, which usually rely on learned distributions or predefined generation procedures, LLMs are better suited to organizing these heterogeneous sources of information during an iterative design process. These considerations motivate an adaptive closed-loop framework in which LLM-based candidate generation is coupled with validation, property evaluation, design memory, and iterative refinement under explicit chemical and structural constraints.

Here, we present LLM-Guided Evolutionary MOF Optimization (LEMO) Agent, a large language model agent framework for the inverse design of MOFs for gas separation. LEMO Agent operates directly on standardized MOFid strings, in which organic fragments, metal nodes, topology labels, and category tags are encoded in a compact symbolic representation~\cite{bucior2019identification}. This text-based representation provides a natural interface between reticular chemistry and LLM-based generation: MOFid strings can be tokenized, edited, compared, and recombined while preserving chemically interpretable information about the building blocks and framework topology. As a result, LLM agents can propose modifications at the level of linkers, metals, and topology labels, and can also relate previous design outcomes to specific symbolic motifs in the generated candidates. In each design round, the agent proposes candidate MOFs according to the target separation objective and the accumulated design context. The generated candidates are then standardized, checked for validity, and evaluated by Transformer property predictors trained for gas separation tasks. The resulting feedback is stored in a structured memory that records successful motifs, failed candidates, recent design history, and reference examples from known MOFs. To balance exploitation and exploration, LEMO Agent further adopts a multi-island strategy~\cite{tanese1989distributed,alba1999survey}, where multiple agents search different regions of chemical space while sharing useful design patterns across iterations. In this way, candidate generation, validation, property evaluation, memory update, and strategy refinement are integrated into a unified design loop rather than treated as isolated steps.

We evaluate LEMO Agent on two representative gas separation tasks, CO$_2$/N$_2$ and CH$_4$/N$_2$~\cite{mason2011evaluating,sumer2017adsorption}, which require different balances between pore geometry, framework chemistry, and gas--framework interactions. For each task, the agent searches the MOFid design space under the same generate--validate--evaluate--remember loop and is compared with non-agentic LLM generation, vanilla agent variants, and conventional generative or optimization baselines. Beyond predicted separation performance, we further assess the generated MOFs in terms of validity, chemical diversity, pore-size reasonableness, and synthetic accessibility, so that the evaluation reflects not only target-property optimization but also the practical constraints of MOF discovery. To further examine whether generated candidates can be translated into physically and experimentally meaningful materials, selected high-performing MOFs are reconstructed from MOFid strings, relaxed, and evaluated by GCMC simulations for binary CO$_2$/N$_2$ and CH$_4$/N$_2$ adsorption. We also establish an experimental down-selection workflow that combines property screening, chemistry-informed feasibility assessment, ligand purchasability checks using PubChem and MolPort, and GPT-assisted synthesis planning based on literature and chemical knowledge. Selected candidates are then subjected to wet-lab synthesis and SEM characterization. The results show that organizing LLM-based generation into a closed-loop agent system improves the efficiency and diversity of MOF inverse design, while the additional simulation and experimental validation steps provide an initial route from symbolic MOF generation toward practical material realization. More broadly, this work demonstrates that LLM agents can serve as adaptive materials design systems that combine symbolic representations, property feedback, chemical knowledge, and design memory to accelerate the discovery of porous materials for gas separation.

\begin{figure}[!htbp]
\centering
\includegraphics[width=0.95\linewidth]{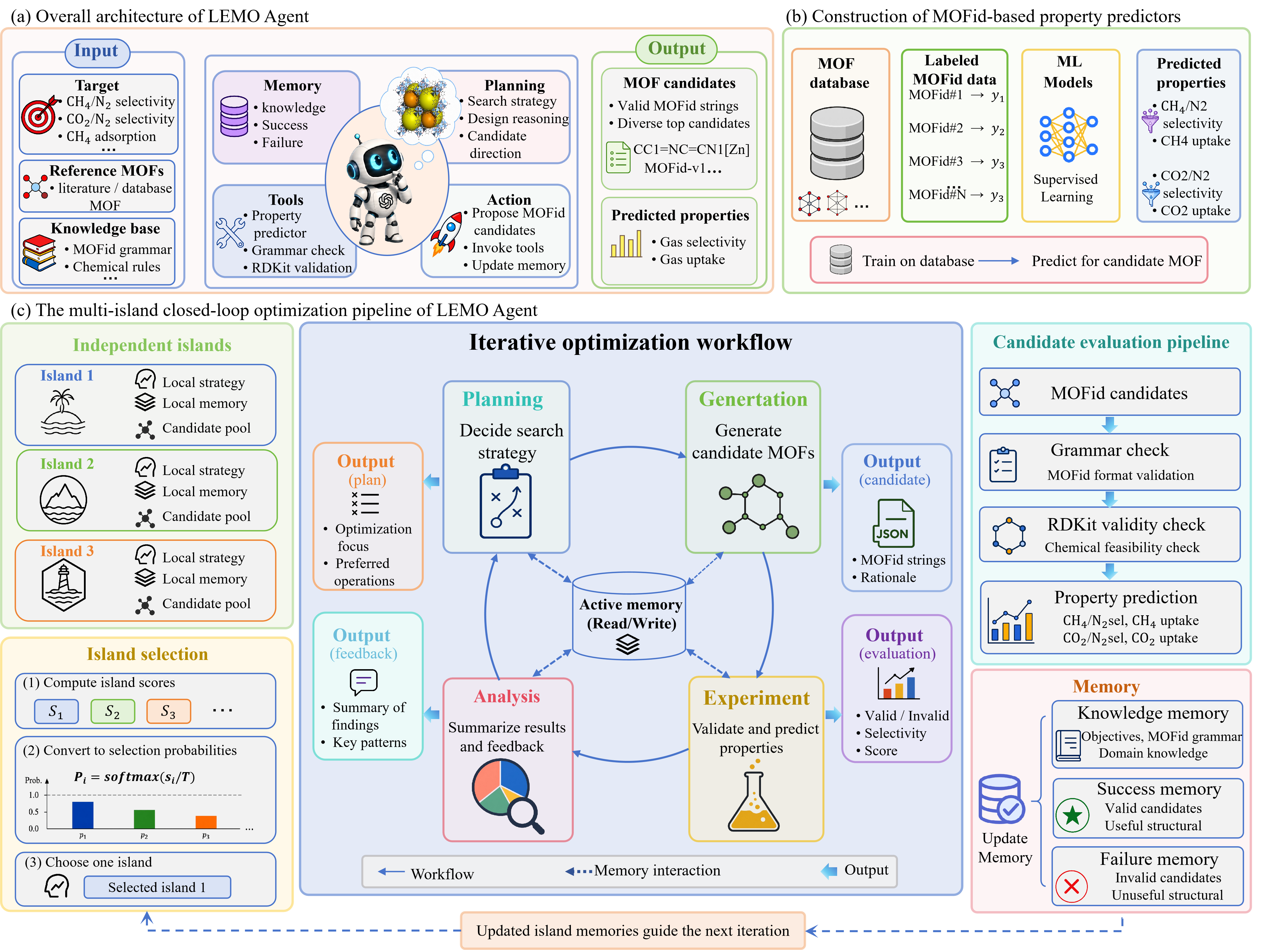}
\caption[Overview of the proposed LEMO Agent]{\textbf{Overview of the proposed LEMO Agent.}
\textbf{a.}~The overall scheme of LEMO Agent: LEMO Agent formulates MOF inverse design as a closed-loop design process in MOFid space. Given a target objective, the agent integrates reference MOFs, domain knowledge, structured memory, validation tools, and property predictors to propose valid MOFid candidates and estimate their target properties. This design enables LEMO Agent to connect MOFid representation, chemically constrained generation, and property-guided candidate discovery.
\textbf{b.}~The construction of MOFid-based property predictors: MOF databases are converted into labeled MOFid data by pairing each MOFid representation with corresponding target properties. Supervised machine-learning models are then trained on these MOFid--property pairs and used to predict gas-separation properties, including CH$_4$/N$_2$ selectivity, CH$_4$ uptake, CO$_2$/N$_2$ selectivity, and CO$_2$ uptake, for newly generated candidate MOFs.
\textbf{c.}~The multi-island closed-loop optimization pipeline of LEMO Agent: LEMO Agent organizes the search into multiple independent islands, each maintaining its own local strategy, local memory, and candidate pool. Island scores are computed and converted into selection probabilities to determine which island is selected for the next optimization step. Within the selected island, the agent iteratively performs planning, generation, candidate evaluation, and analysis while actively reading from and writing to memory. Generated candidates are evaluated through MOFid grammar checking, RDKit validity checking, and property prediction. The resulting valid candidates, high-performing candidates, and failed cases are summarized into knowledge, success, and failure memories, which are then used to guide subsequent optimization rounds.}
\label{fig:overview}
\end{figure}

\section{Results}\label{sec2}
\subsection{Overview of LEMO Agent}
We developed LEMO Agent, a closed-loop language-agent framework for the inverse design of metal--organic frameworks in MOFid space. The central goal of LEMO Agent is to transform a general-purpose large language model from a one-shot sequence generator into an iterative materials design agent that can propose, evaluate, remember, and refine candidate MOFs under explicit chemical and task-specific constraints. Instead of directly generating crystallographic structures, LEMO Agent operates on MOFid representations, which encode organic fragments, metal-containing building units, and topology information in a compact string format. This representation provides a practical interface for language-based generation while retaining chemically meaningful information required for validation and property prediction.

The overall scheme of LEMO Agent is shown in Fig~\ref{fig:overview} a. Given a target gas-separation objective, such as high CH$_4$/N$_2$ selectivity, CO$_2$/N$_2$ selectivity, the agent integrates reference MOFs, domain knowledge, structured memory, validation tools, and property predictors to guide candidate generation. At each iteration, the planner analyzes the current design context, including the target objective, MOFid grammar, chemical rules, prior successful candidates, and failed cases, and then formulates a search strategy for the next round. The generator subsequently proposes new MOFid candidates under these constraints, while the action module invokes validation and prediction tools and updates the memory according to the evaluation results.

To provide efficient property feedback during optimization, we first construct MOFid-based property predictors from labeled MOF data, as illustrated in Fig~\ref{fig:overview} b. MOF databases are converted into standardized MOFid representations and paired with target properties, such as CH$_4$/N$_2$ selectivity, CH$_4$ uptake, CO$_2$/N$_2$ selectivity, and CO$_2$ uptake. Supervised machine-learning models are then trained on these MOFid--property pairs and used as fast surrogate evaluators for newly generated candidates. This enables LEMO Agent to rapidly estimate gas-separation performance without performing expensive simulations or experiments for every generated MOF.

LEMO Agent further organizes the search using a multi-island closed-loop optimization pipeline, as shown in Fig~\ref{fig:overview} c. Each island maintains its own local strategy, local memory, and candidate pool, allowing different regions of MOFid space to be explored in parallel. During optimization, island scores are computed from the accumulated performance of each island and converted into selection probabilities, which determine which island is selected for the next generation step. This design helps balance exploitation of high-performing design patterns with exploration of diverse linker, metal-node, and topology combinations.

Within the selected island, LEMO Agent iteratively executes planning, generation, validation and property prediction, and analysis while actively reading from and writing to memory. The analysis module summarizes the evaluation outcomes, including valid high-performing candidates and invalid or low-performing cases, and extracts useful patterns for subsequent optimization. The memory module stores different types of design evidence in separated components, including knowledge memory, success memory, and failure memory. Knowledge memory preserves objectives, MOFid grammar, and domain knowledge; success memory records valid candidates and useful structural patterns; and failure memory stores invalid candidates and unfavorable structural patterns. These memories provide feedback to the planner in later iterations, enabling the agent to learn not only from successful designs but also from repeated failure modes.

Through this closed-loop design, LEMO Agent differs from both conventional high-throughput screening and standard language-model generation. High-throughput screening evaluates a fixed candidate library and is limited by the structures already enumerated, whereas LEMO Agent actively proposes new MOF candidates during the search. A vanilla language model can generate MOFid-like strings, but it does not automatically enforce MOFid grammar, verify chemical validity, incorporate property feedback, or remember failure patterns across iterations. In contrast, LEMO Agent integrates language-based proposal, rule-based validation, surrogate property prediction, structured memory, and multi-island optimization into a unified workflow. This enables chemically constrained exploration of MOF space and provides an efficient route toward valid, diverse, and high-performing MOF candidates.

\subsection{Transformer-based MOFid predictors for gas-separation evaluation}

To provide rapid and task-specific feedback during the closed-loop optimization of LEMO Agent, we constructed Transformer-based~\cite{vaswani2017attention} property predictors that directly operate on MOFid representations. As shown in Fig.~2a, each MOF structure is represented by an MOFid string that encodes key reticular components, including organic linkers, metal nodes, topology information, and category labels. This string-based representation enables the predictor to learn structure--property relationships without explicitly reconstructing the full crystallographic structure for every generated candidate, making it suitable for efficient evaluation during iterative MOF design.

The predictor takes a cleaned MOFid string as input and tokenizes it into a sequence containing chemically and structurally meaningful elements, such as linker fragments, metal-containing units, topology identifiers, and category tokens. The tokenized sequence is then processed by a Transformer encoder, where self-attention layers capture contextual dependencies among different MOFid components. The encoded representation is aggregated through the special classification token and passed to regression heads for predicting gas-separation-related properties. In this work, we trained task-specific predictors for four objectives: CH$_4$/N$_2$ selectivity, CH$_4$ adsorption capacity, CO$_2$/N$_2$ selectivity, and CO$_2$ adsorption capacity.

Before model training, MOFid records were standardized and filtered to ensure that the predictors were trained on valid and consistent inputs. Malformed MOFid strings, unknown fragments, wildcard atoms, empty fields, duplicated entries, and invalid category labels were removed. The organic fragment prefix of each MOFid was further checked using cheminformatics tools to exclude chemically invalid or non-parseable candidates.

The predictive performance was evaluated on held-out test sets by comparing the simulated properties with the model predictions. As shown in Fig.~2b--e, the Transformer-based MOFid predictors achieved strong agreement with the simulated values across all four gas-separation tasks. For CH$_4$/N$_2$ selectivity, the predictor achieved an $R^2$ of 0.94 and a mean absolute error (MAE) of 0.26. For CH$_4$ adsorption capacity, the model achieved an $R^2$ of 0.93 and an MAE of 0.07. The CO$_2$/N$_2$ selectivity predictor reached an $R^2$ of 0.84 with an MAE of 7.78, while the CO$_2$ adsorption capacity predictor achieved an $R^2$ of 0.88 and an MAE of 0.15. These results indicate that MOFid strings retain sufficient structural information for learning gas-separation-relevant properties.

\begin{figure*}[!htbp]
\centering
\includegraphics[width=0.98\textwidth]{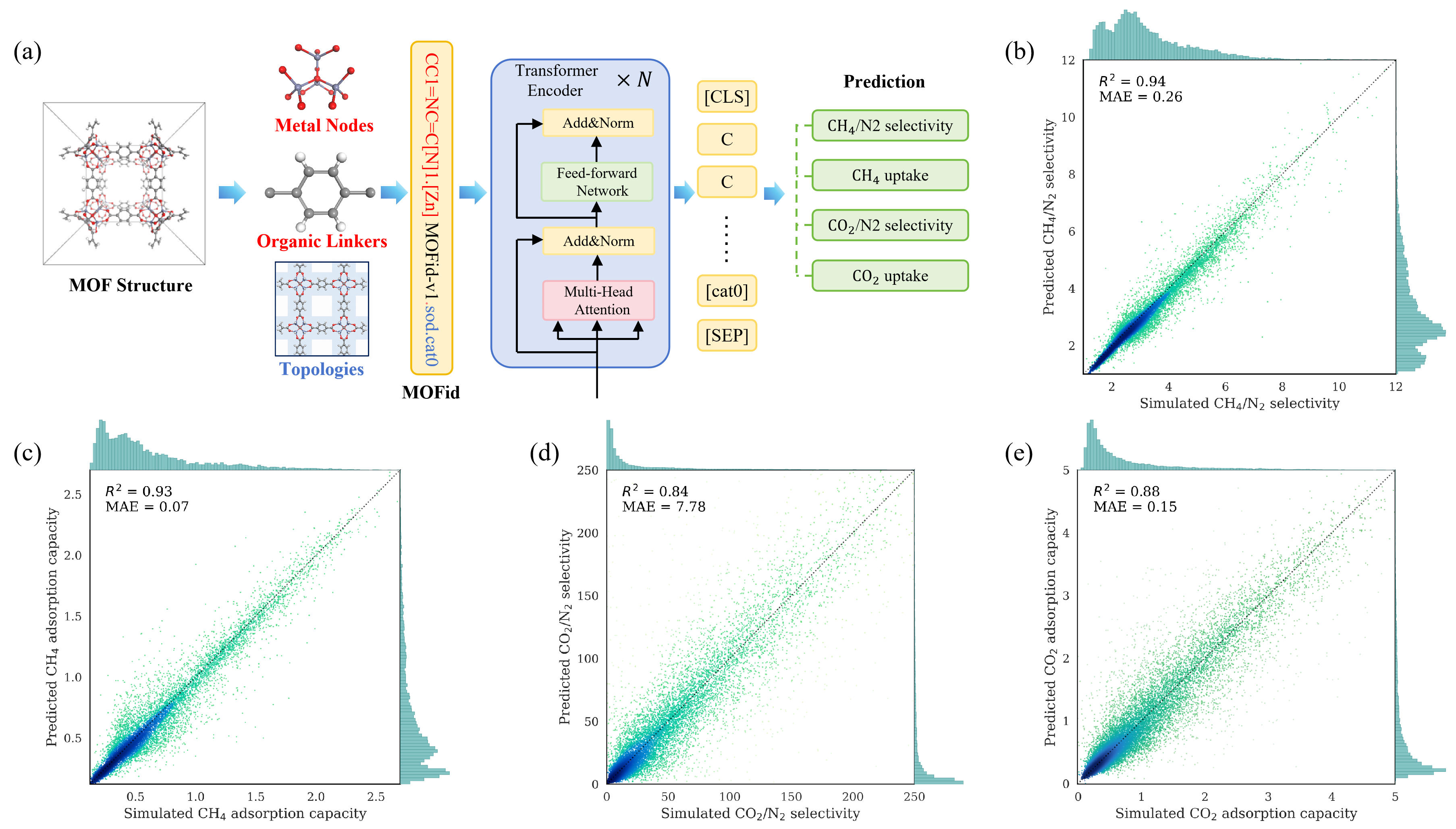}
\caption[Transformer-based MOFid predictors for gas-separation evaluation]{\textbf{Transformer-based MOFid predictors for gas-separation evaluation.}
\textbf{a.}~Schematic illustration of the MOFid-based property prediction model. Each MOF structure is converted into an MOFid string that encodes metal nodes, organic linkers, topology information, and category labels. The tokenized MOFid sequence is processed by a Transformer encoder to predict gas-separation-related properties, including CH$_4$/N$_2$ selectivity, CH$_4$ adsorption capacity, CO$_2$/N$_2$ selectivity, and CO$_2$ adsorption capacity.
\textbf{b--e.}~Prediction performance of the Transformer-based MOFid predictors on held-out test sets for the four gas-separation tasks. The predicted values show strong agreement with the simulated values, with $R^2$ values of 0.94, 0.93, 0.84, and 0.88 for CH$_4$/N$_2$ selectivity, CH$_4$ adsorption capacity, CO$_2$/N$_2$ selectivity, and CO$_2$ adsorption capacity, respectively. The dashed diagonal lines indicate perfect prediction.}
\label{fig}
\end{figure*}

\subsection{LEMO Agent for gas-separation MOF design}\label{sec:results_property}

We evaluated LEMO Agent on gas-separation MOF design directly in MOF space. Two representative separation tasks were considered, CH$_4$/N$_2$ and CO$_2$/N$_2$ separation, to examine whether the proposed agent can generate valid MOF candidates with improved predicted separation performance. To assess the contribution of the proposed agentic optimization strategy, we compared LEMO Agent with several baselines, including an ablated variant without the multi-island mechanism, denoted as LEMO w/o MI, a vanilla agent, an LLM-only baseline, and MOFGPT~\cite{badrinarayanan2025mofgpt}, a reinforcement-learning-based generative MOF design method.

\begin{figure*}[t]
    \centering 
    \includegraphics[width=0.97\textwidth]{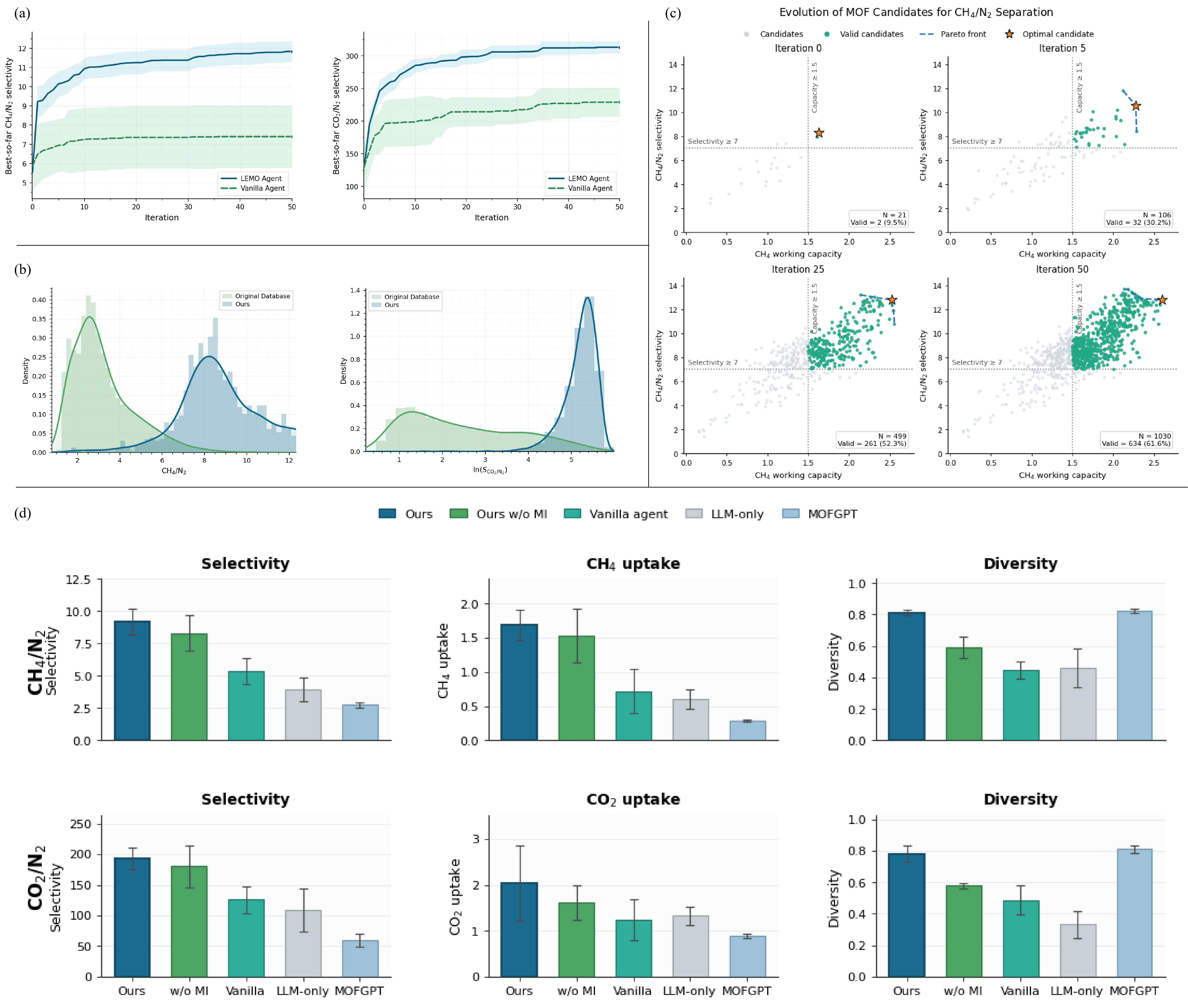} 
    \caption{\textbf{LEMO Agent for gas-separation MOF design.}
    \textbf{a,} Best-so-far predicted selectivity over 50 iterations for CH$_4$/N$_2$ and CO$_2$/N$_2$ separation. LEMO Agent continuously improves the best generated candidates and outperforms the vanilla agent on both tasks.
    \textbf{b,} Property distributions of the original database and LEMO Agent-generated candidates. The generated candidates are shifted toward higher CH$_4$/N$_2$ selectivity and higher log-transformed CO$_2$/N$_2$ selectivity, indicating enrichment of high-performance MOFs.
    \textbf{c,} Evolution of generated candidates for CH$_4$/N$_2$ separation in the CH$_4$ working capacity--selectivity space. Gray points denote all generated candidates, green points denote valid candidates, blue dashed lines indicate the Pareto front, and orange stars indicate the optimal candidate at each stage. The dashed threshold lines indicate CH$_4$/N$_2$ selectivity $\geq 7$ and CH$_4$ working capacity $\geq 1.5$.
    \textbf{d,} Quantitative comparison of LEMO Agent with ablated and baseline methods on CH$_4$/N$_2$ and CO$_2$/N$_2$ design tasks. LEMO w/o MI denotes the ablated variant without the multi-island mechanism. The metrics include selectivity, working capacity, and structural diversity. Bars represent mean values and error bars indicate standard deviations across independent runs.}
    \label{fig:lemo_gas_design}
\end{figure*}

As shown in Fig.~\ref{fig:lemo_gas_design}a, LEMO Agent consistently improves the best-so-far predicted selectivity during the optimization process. For CH$_4$/N$_2$ separation, the best generated candidate increases rapidly in the early iterations and continues to improve gradually until convergence, reaching a substantially higher selectivity than the vanilla agent. In contrast, the vanilla agent shows only limited improvement and quickly saturates at a lower level. A similar trend is observed for CO$_2$/N$_2$ separation, where LEMO Agent rapidly identifies high-selectivity candidates and maintains a clear advantage throughout the 50-iteration optimization process. These results indicate that the feedback-driven agentic design strategy can effectively guide MOF generation toward improved gas-separation performance.

We further compared the property distributions of generated candidates with those of the original database. For CH$_4$/N$_2$ separation, the original database is mainly concentrated in the low-selectivity region, whereas the LEMO Agent-generated candidates are shifted toward substantially higher predicted selectivity values (Fig.~\ref{fig:lemo_gas_design}b). For CO$_2$/N$_2$ separation, the log-transformed selectivity distribution shows a similar trend: generated candidates are strongly enriched in the high-selectivity region compared with the original database. These distributional shifts suggest that LEMO Agent does not simply reproduce the training database distribution, but actively explores MOF regions with improved predicted gas-separation properties.

To analyze the optimization trajectory in more detail, we visualized the CH$_4$/N$_2$ design process in the two-dimensional space of CH$_4$ working capacity and CH$_4$/N$_2$ selectivity (Fig.~\ref{fig:lemo_gas_design}c). At the initial stage, only a small fraction of generated candidates are valid, and most candidates remain outside the high-performance region. As optimization proceeds, the fraction of valid candidates increases from 9.5\% at iteration 0 to 30.2\%, 52.3\%, and 61.6\% at iterations 10, 25, and 50, respectively. Meanwhile, valid candidates increasingly populate the region satisfying both the selectivity and working-capacity thresholds. The Pareto front~\cite{miettinen1999nonlinear} also shifts toward the upper-right region, indicating simultaneous improvement in CH$_4$/N$_2$ selectivity and CH$_4$ working capacity. This trajectory demonstrates that LEMO Agent improves not only the best individual candidate, but also the overall feasibility and multi-objective quality of the generated MOF population.

We then quantitatively benchmarked LEMO Agent against ablated and baseline methods on both gas-separation tasks (Fig.~\ref{fig:lemo_gas_design}d). Across CH$_4$/N$_2$ and CO$_2$/N$_2$ separation, LEMO Agent achieves the best overall property performance, giving the highest selectivity and working capacity among all compared methods. This result shows that LEMO Agent can generate MOFids that not only satisfy syntactic and validity requirements, but also move toward the target gas-separation objectives. In comparison, the vanilla agent and LLM-only baseline show much weaker property optimization, suggesting that direct LLM generation or weakly guided generation is insufficient for efficiently searching the constrained MOF design space.

An important advantage of LEMO Agent is that its property improvement is not obtained at the cost of severe diversity loss. As shown in Fig.~\ref{fig:lemo_gas_design}d, LEMO Agent maintains high structural diversity on both tasks, with diversity comparable to MOFGPT. This comparison is particularly informative because MOFGPT is a reinforcement-learning-based generative MOF design method and can generate diverse MOF-like candidates. However, although MOFGPT achieves high diversity, its selectivity and working capacity remain substantially lower than those of LEMO Agent. These results indicate that diversity alone is not sufficient for target-oriented MOF design. LEMO Agent achieves a more favorable balance by combining generation, validation, property feedback, and iterative refinement, thereby preserving broad structural exploration while enriching high-performing candidates.

The comparison with LLM-based baselines further highlights the necessity of explicit exploration control. The vanilla agent and LLM-only baseline exhibit much lower diversity than the LEMO Agent, in some cases approaching only about half of the diversity achieved by the proposed method. This suggests that conventional LLM-based generation tends to repeatedly sample a limited set of familiar linker, metal-node, or topology patterns. Such structural concentration restricts the exploration of MOF space and limits the chance of discovering high-performing candidates. Therefore, the lower performance of these baselines is closely associated with their restricted structural coverage.

The ablation study confirms that the multi-island mechanism is a key component for maintaining this performance--diversity balance. After removing the multi-island mechanism, both property performance and structural diversity decrease. The drop in diversity indicates that a single optimization trajectory is more likely to over-exploit locally successful structural patterns, while losing access to alternative linker, metal, and topology combinations. Meanwhile, the decrease in selectivity and working capacity suggests that reduced exploration also weakens the ability to find better gas-separation candidates. Therefore, the multi-island mechanism does not merely increase the number of generated samples; it helps organize the search into multiple complementary directions, preventing premature convergence and improving the robustness of closed-loop MOF optimization.

Overall, Fig.~\ref{fig:lemo_gas_design} shows that LEMO Agent effectively optimizes gas-separation MOFs. Across iterations, the framework steadily improves predicted performance, enriches high-selectivity candidates, and expands the Pareto front while maintaining chemical diversity.

\subsection{Physical reasonableness, synthetic accessibility, and chemical diversity of generated MOFs}

Beyond optimizing gas-separation performance, a practical MOF design method should generate candidates that are not only high-scoring in the target objective but also physically meaningful, synthetically plausible, and chemically diverse. We therefore evaluated the generated MOFids from three complementary perspectives: MOFid-predicted physical properties, linker-level synthetic accessibility, and chemical diversity across organic linkers and metal nodes.

We first examined the physical reasonableness of the generated MOFs using two structural descriptors predicted directly from MOFid representations: pore limiting diameter (PLD) and framework density $\rho$. Following the method of Situ et al~\cite{situ2025efficient}., PLD and density were estimated using machine-learning surrogate models based on MOFid-derived representations, rather than being directly calculated from fully reconstructed and geometry-optimized CIF structures. PLD is closely related to molecular accessibility and size-dependent transport through the framework, making it particularly relevant for gas-separation applications. Framework density provides a complementary global descriptor of structural compactness and helps assess whether the generated candidates correspond to realistic crystalline frameworks rather than unrealistically open or highly sparse networks. The details of these MOFid-based prediction models and their validation are provided in Supplementary Section A.

As shown in Fig.~\ref{fig:physical_synthetic_diversity}a, the predicted PLD and density distributions of the generated MOFs are generally comparable to those of literature gas-separation MOFs. In contrast, ARC-MOF spans a broader structural space, including structures with larger pores and lower framework densities. These results suggest that the generated MOFids remain within physically reasonable regions of MOF structural space, rather than relying on excessively open or sparse frameworks to improve predicted performance.

We next evaluated the synthetic feasibility of the organic linkers extracted from the generated MOFids. Specifically, we calculated the synthetic accessibility score (SA score) and the synthetic complexity score (SCScore)~\cite{ertl2009estimation,coley2018scscore} for the generated organic fragments. As shown in Fig.~\ref{fig:physical_synthetic_diversity}b, most generated linkers fall within the range of readily accessible or moderately complex molecules. Only a small fraction of generated linkers exhibit relatively high synthetic complexity. The consistent trends observed from both SA score and SCScore indicate that LEMO does not rely on overly complicated organic fragments to achieve high predicted separation performance. Instead, the closed-loop optimization process can improve target properties while largely preserving linker-level synthetic accessibility.

We finally investigated the chemical diversity of the generated MOFs from both organic and inorganic components. For the organic component, Fig.~\ref{fig:physical_synthetic_diversity}c visualizes the linker chemical space using t-SNE. The generated linkers occupy a broad region within the ARC-MOF chemical space and partially overlap with literature MOF linkers, indicating that LEMO captures chemically meaningful linker patterns from known MOFs. At the same time, many generated linkers are distributed outside the most densely populated literature regions, suggesting that the model can explore new linker combinations rather than simply reproducing existing examples.

For the inorganic component, we analyzed the metal-element distribution of the generated MOFs. As shown in Fig.~\ref{fig:physical_synthetic_diversity}d, the generated candidates cover a diverse set of metal elements, including commonly used MOF metals such as Zn, Cu, Ni, and Al, as well as several less frequently explored elements. Compared with the literature set, which is more concentrated around a limited number of commonly reported metals, the generated MOFs exhibit a broader metal distribution while still remaining within chemically reasonable metal choices. This behavior is consistent with our design strategy, which encourages the exploration of diverse metal nodes while keeping the inorganic fragments simple and MOFid-parseable.

Overall, these analyses demonstrate that LEMO generates high-performing MOF candidates without sacrificing physical plausibility, linker-level synthetic accessibility, or chemical diversity. The generated MOFs show predicted PLD and density distributions comparable to reported gas-separation MOFs, while simultaneously expanding the explored chemical space of both organic linkers and metal nodes.

\begin{figure*}[t]
    \centering
    \includegraphics[width=\textwidth]{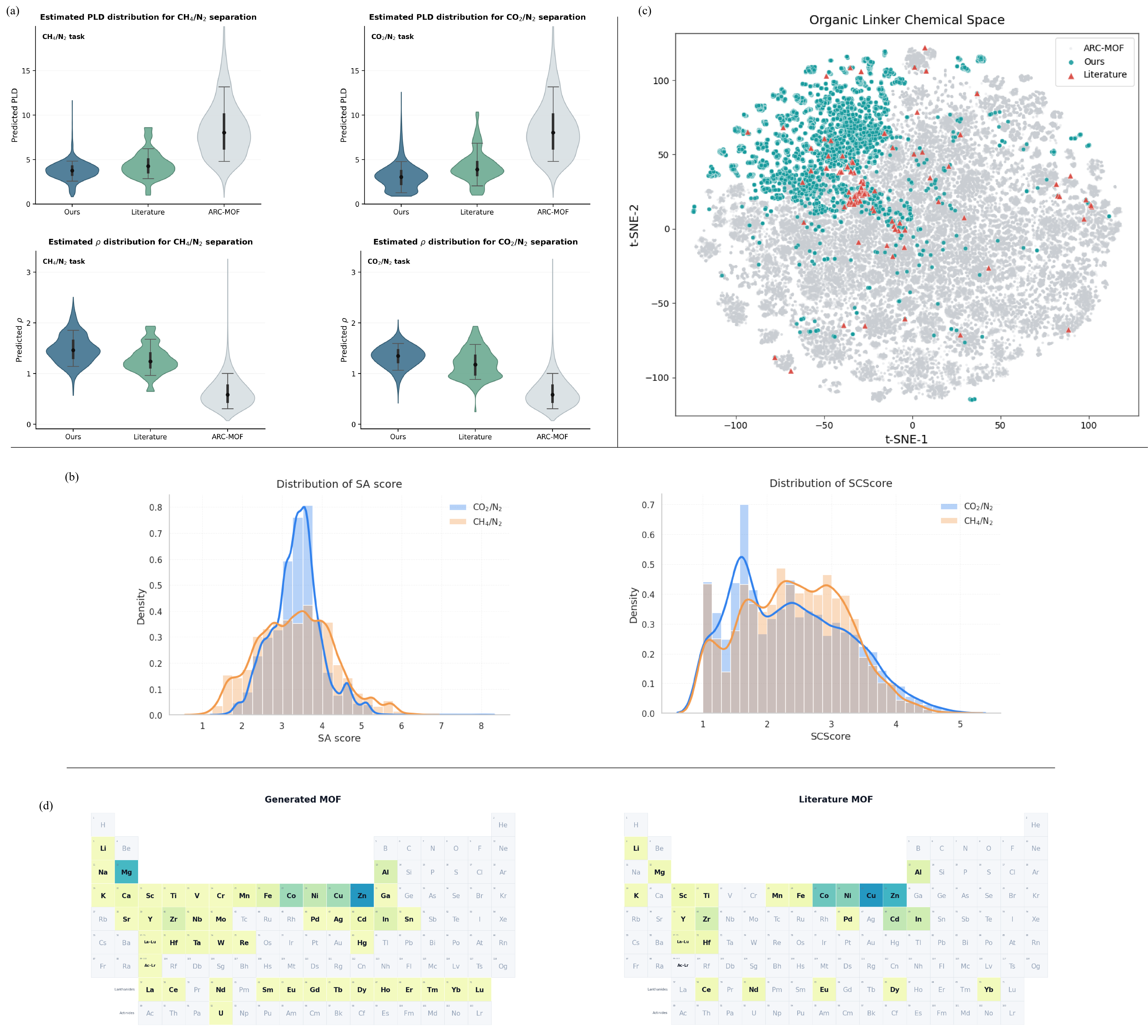}
    \caption{
    Physical reasonableness, synthetic accessibility, and chemical diversity of generated MOFs.
    \textbf{a}, MOFid-predicted physical properties of generated MOFs, literature MOFs, and ARC-MOF structures, including pore limiting diameter (PLD) and framework density for the CH$_4$/N$_2$ and CO$_2$/N$_2$ separation tasks. PLD and density were estimated from MOFid representations using machine-learning surrogate models, as described in the Supplementary Section A.
    \textbf{b}, Distributions of synthetic accessibility score (SA score) and synthetic complexity score (SCScore) for organic linkers extracted from generated MOFids in the CO$_2$/N$_2$ and CH$_4$/N$_2$ tasks.
    \textbf{c}, t-SNE visualization of organic-linker chemical space. Gray points represent ARC-MOF linkers, cyan points represent generated linkers, and red triangles represent literature MOF linkers.
    \textbf{d}, Periodic-table heat maps showing the metal-element distributions in generated MOFs and literature MOFs.
    }
    \label{fig:physical_synthetic_diversity}
\end{figure*}

\subsection{Validation of MOFs Material}

\label{sec:validation_mofs}

To further validate the practical relevance of the MOFs generated by LEMO Agent, we combined molecular-simulation-based evaluation with an experimental screening and synthesis workflow. Representative high-performing candidates were first reconstructed from MOFid strings into crystallographic structures. The generated MOFids were converted into CIF files using PORMAKE, followed by geometry relaxation with the MACE-MP-0 machine-learned potential through the ASE interface. The optimized structures were then evaluated by grand canonical Monte Carlo (GCMC) simulations in RASPA for binary gas adsorption. CO$_2$/N$_2$ adsorption was simulated at 298 K and 100 kPa using a 15:85 gas mixture, while CH$_4$/N$_2$ adsorption was simulated at 298 K and 100 kPa using a 50:50 gas mixture. Framework atoms were described using UFF parameters with EQeq charges, and gas molecules were modeled using the TraPPE force field. The adsorption selectivity was calculated from the simulated mixture loadings as
\begin{equation*}
S_{A/B}=\frac{q_A/q_B}{y_A/y_B},
\end{equation*}
where $q_i$ and $y_i$ denote the adsorbed loading and gas-phase mole fraction of component $i$, respectively.

\begin{figure*}[t]
    \centering
    \includegraphics[width=\textwidth]{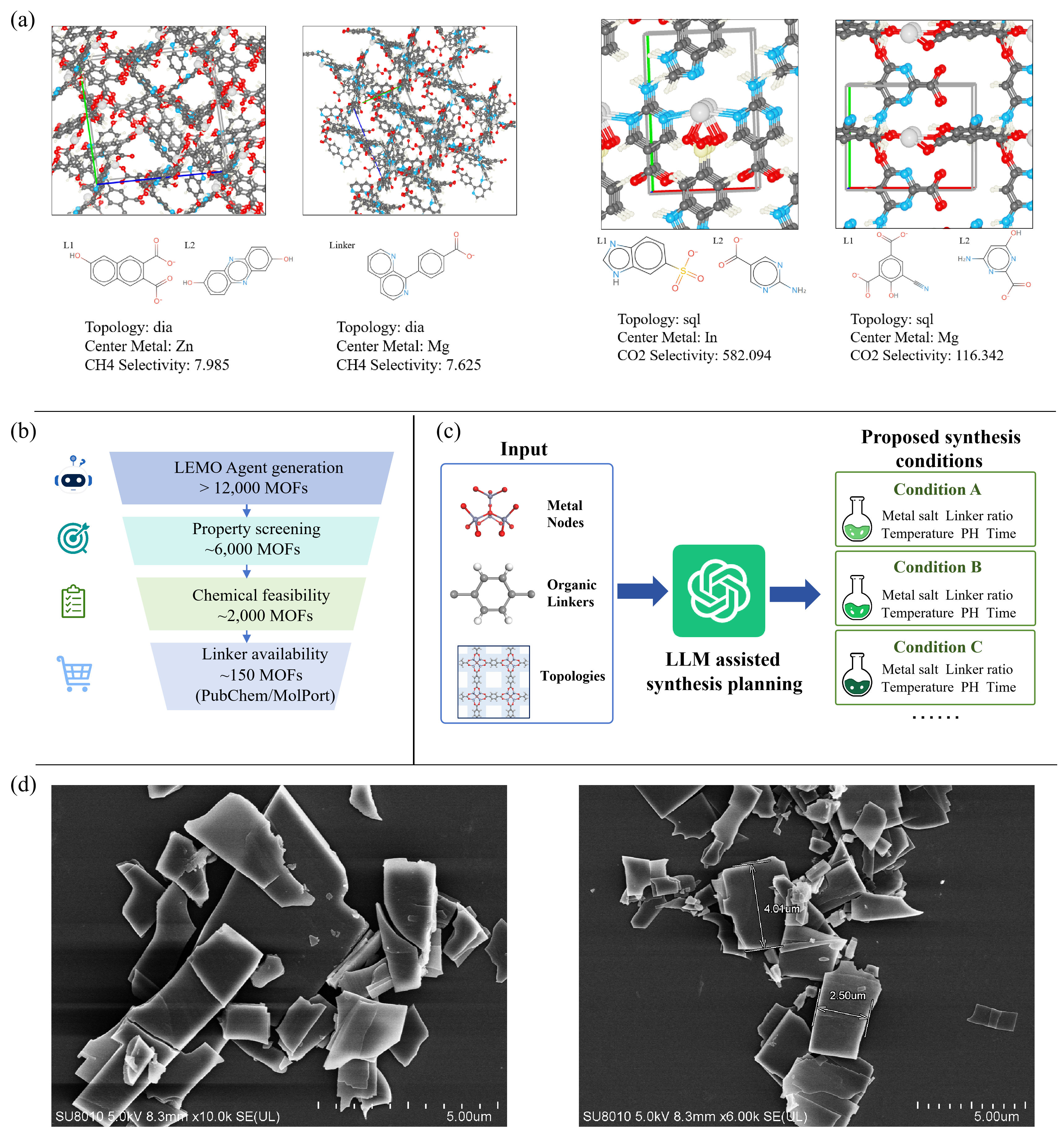}
    \caption{
    Validation of generated MOF materials. 
    (a) Representative LEMO-generated MOFs reconstructed from MOFid strings and evaluated by GCMC simulations for CH$_4$/N$_2$ and CO$_2$/N$_2$ separation. The framework structures, linker compositions, topology, center metal, and simulated selectivity are shown for selected candidates. 
    (b) Experimental down-selection and synthesis-planning workflow. More than 12,000 generated MOFs were first screened by predicted separation performance, then filtered by chemistry-informed feasibility rules and ligand commercial availability using PubChem and MolPort. Shortlisted candidates were provided to GPT to recommend synthesis conditions from literature and chemical knowledge.
    (c) SEM images of the synthesized product, showing plate-like crystalline particles on the micrometer scale.
    }
    \label{fig:validation_mofs}
\end{figure*}

As shown in Fig.~\ref{fig:validation_mofs}a, several generated MOFs exhibit promising simulated separation performance for CH$_4$/N$_2$ or CO$_2$/N$_2$, indicating that the LEMO-generated candidates are not only valid in MOFid space but can also be reconstructed into physically meaningful framework structures for adsorption simulation. These GCMC results provide a higher-fidelity validation of the surrogate-predicted candidates and support the ability of the agent to identify MOFs with favorable gas-separation behavior.

We then designed an experimental down-selection workflow to identify candidates that are more suitable for wet-lab synthesis. Starting from more than 12,000 generated MOFs, we first retained approximately 6,000 candidates with favorable predicted selectivity and uptake-related performance. The remaining candidates were further filtered by chemistry-informed feasibility rules derived from coordination chemistry, including coordination strength, framework-forming tendency, and mixed-linker compatibility. In particular, we prioritized MOFs containing multidentate carboxylate and N-donor linkers that are more likely to form extended frameworks, while deprioritizing combinations dominated by weakly coordinating or effectively monodentate motifs. This step reduced the pool to approximately 2,000 candidates. We then screened the organic linkers for commercial purchasability using PubChem and MolPort, yielding approximately 150 candidates with purchasable ligands.

For the final shortlist, the MOF composition, linker structures, metal nodes, and topology information were provided to GPT to generate synthesis recommendations based on reported MOF literature and chemical knowledge. The SEM images in Fig.~\ref{fig:validation_mofs}c show plate-like crystalline particles with micrometer-scale lateral sizes, providing initial experimental evidence that the LEMO Agent to synthesis workflow can lead to realistic MOF materials.

\section{Discussion}
\label{sec3}

In this work, we introduced LEMO Agent, a chemically constrained and memory-guided language-agent framework for property-guided inverse design of metal--organic frameworks. By integrating language-based MOFid generation, explicit validity checking, Transformer-based property prediction, structured memory, and multi-island exploration, LEMO Agent enables iterative optimization directly in MOF space. Across CH$_4$/N$_2$ and CO$_2$/N$_2$ separation tasks, the framework improves the discovery of valid and high-performing candidates over one-shot language generation and ablated agent variants, while preserving novelty, chemical diversity, physical plausibility and linker-level synthetic accessibility.

A central advantage of LEMO Agent comes from the use of MOFid as the design representation. MOFid encodes organic fragments, metal-containing units and topology in a compact textual form, making it naturally compatible with large language models. This allows the agent to make use of the text-based reasoning ability and chemical pattern knowledge of LLMs for MOF design. Rather than treating MOF generation as a fixed combinatorial search problem, LEMO Agent can reason over linker chemistry, metal nodes, topology choices and previous design feedback within a unified language-based workflow.

This representation also distinguishes LEMO Agent from conventional optimization strategies such as genetic algorithms. Genetic algorithms are often confined to a predefined building-block set, meaning that the accessible chemical space is largely determined before optimization begins. The algorithm can only recombine, mutate or select from fragments already provided by the user. In contrast, LEMO Agent can propose new MOFid strings by leveraging the chemical knowledge encoded in the language model, while explicit MOFid validation, property prediction and memory-guided feedback constrain the generation toward chemically meaningful and task-relevant candidates. This enables the framework to explore new combinations of organic linkers, metal nodes and topologies beyond a fixed building-block library.

The role of the language model in LEMO Agent is therefore not limited to sequence generation. It serves as an adaptive proposal engine within a closed-loop scientific design system, where candidate generation is separated from chemical validation, property evaluation and memory update. This modular design makes the search process interpretable and readily adaptable to different design objectives. The improvement over non-iterative and memory-ablated baselines suggests that the performance gain arises from the coordinated use of planning, feedback, structured memory and diversity-preserving exploration, rather than from repeated random sampling or scalar score optimization alone.

Structured memory is an important component of LEMO Agent. Instead of storing all previous generations as an undifferentiated history, the framework organizes information into reference examples, successful candidates, failed patterns and promising frontier structures. This helps the planner reuse useful motifs, avoid recurring invalid designs and refine near-threshold candidates. The multi-island strategy further maintains multiple memory pools to explore different regions of MOF space, reducing premature convergence and improving diversity.

The generated candidates were evaluated beyond predicted gas-separation performance. We assessed them from multiple perspectives, including physical properties, linker synthetic accessibility, chemical diversity, and higher-fidelity adsorption simulation. The generated MOFs show pore-size distributions and density ranges comparable to existing gas-separation MOFs, while their linkers generally exhibit moderate synthetic complexity. Selected high-performing candidates were further reconstructed from MOFid strings, relaxed, and evaluated by GCMC simulations for binary CH$_4$/N$_2$ and CO$_2$/N$_2$ adsorption. These simulations provide an additional validation step beyond surrogate prediction and show that representative LEMO-generated candidates can be converted into physically meaningful framework structures with promising simulated separation behavior.

We also explored the experimental actionability of the generated MOFs through a down-selection and synthesis-planning workflow. Starting from the generated candidate pool, we first selected MOFs with favorable predicted selectivity and uptake-related performance, then applied chemistry-informed feasibility screening based on coordination strength, framework-forming tendency, and mixed-linker compatibility. The remaining candidates were further filtered by checking whether their organic linkers could be purchased from PubChem- and MolPort-indexed suppliers. For the final shortlist, GPT was used to recommend synthesis conditions based on reported MOF literature and chemical knowledge. Selected candidates were then subjected to wet-lab synthesis and SEM characterization. Although these experiments represent an initial validation rather than a complete synthesis-performance study, they demonstrate a practical route for translating symbolic MOF generation into experimentally testable materials.

Overall, LEMO Agent demonstrates that chemically constrained language agents can provide an effective route for inverse MOF design. By combining the textual compatibility of MOFid, the chemical knowledge encoded in LLMs, and closed-loop validation, prediction, and memory update, the framework enables efficient exploration of new MOF candidates beyond predefined building-block libraries. More broadly, this design principle can be extended to other reticular materials and molecular systems where compact textual representations, property predictors, and validation rules are available. LEMO Agent thus represents a step toward interpretable, memory-guided, and chemically constrained language-agent systems for autonomous materials discovery.

\section{Methods}
\subsection{Datasets}
\label{sec:methods_datasets}

The data used in this study were collected from multiple sources, including large-scale computational MOF datasets and experimentally reported gas-separation MOFs. For the CO$_2$/N$_2$ separation task, we used the post-combustion vacuum swing adsorption (VSA) data from the ARC-MOF dataset, which contains approximately 220,000 computationally characterized MOFs with adsorption and separation related properties. For the CH$_4$/N$_2$ separation task, we used the simulated adsorption and separation dataset reported by Li et al., which contains approximately 110,000 MOF structures with corresponding CH$_4$/N$_2$ performance labels. These two computational datasets served as the main data sources for training task-specific property predictors and defining the optimization objectives of LEMO Agent.

In addition to these large-scale computational datasets, we collected approximately 200 experimentally synthesized MOFs from the literature for CH$_4$/N$_2$ and CO$_2$/N$_2$ separations. This literature set was used both as initial examples and domain knowledge for guiding the agent, and as an experimentally grounded reference for downstream analysis, including comparisons of MOFid-predicted physical properties, linker-level synthetic accessibility, and chemical diversity.

\subsection{Data preprocessing and MOFid representation}
\label{sec:methods_preprocessing}

All MOF structures were represented using MOFid-v1 strings before model training, candidate generation, and evaluation. For MOFs with available crystallographic information files (CIFs), we first used the MOFid toolkit to convert CIF structures into MOFid-v1 representations. 

After MOFid conversion, we removed entries with abnormal or incomplete MOFid outputs. In particular, structures were excluded if the topology field was labelled as \texttt{unknown}, \texttt{error}, empty, or otherwise failed to provide a valid framework topology. Entries with missing MOFid strings, incomplete \texttt{MOFid-v1} suffixes, or unsupported cat labels were also removed. 

Before model training, candidate generation, and evaluation, all MOFid strings were standardized into a unified format. We required the organic fragment(s) to appear first, followed by the metal or metal-cluster fragment. This standardized representation was used consistently across the property-prediction models, LEMO Agent generation prompts, validation pipeline, and downstream analyses.

\subsection{Transformer-based MOFid property predictors}
\label{sec:methods_predictors}

To provide fast property feedback during closed-loop optimization, we trained Transformer-based regression models to predict gas-separation properties directly from standardized MOFid strings. Each MOFid was tokenized into a sequence of discrete tokens, including organic fragments, metal-containing fragments, topology labels, cat values, and separator tokens. The token sequence was converted into continuous embeddings and used as the input to a Transformer encoder.

In each Transformer encoder layer, self-attention was used to capture long-range dependencies among organic fragments, metal nodes, and topology tokens. Given the hidden representation $\mathbf{H}$ of a MOFid sequence, the query, key, and value matrices were computed as
\begin{equation}
\mathbf{Q}=\mathbf{H}\mathbf{W}^{Q}, \quad
\mathbf{K}=\mathbf{H}\mathbf{W}^{K}, \quad
\mathbf{V}=\mathbf{H}\mathbf{W}^{V},
\end{equation}
where $\mathbf{W}^{Q}$, $\mathbf{W}^{K}$, and $\mathbf{W}^{V}$ are learnable projection matrices. The self-attention output was calculated by scaled dot-product attention:
\begin{equation}
\mathrm{Attention}(\mathbf{Q},\mathbf{K},\mathbf{V})
=
\mathrm{softmax}
\left(
\frac{\mathbf{Q}\mathbf{K}^{\top}}{\sqrt{d}}
\right)
\mathbf{V},
\end{equation}
where $d$ is the hidden dimension of each attention head. The final sequence representation was passed through a regression head to predict the target property.

Separate predictors were trained for different gas-separation properties, including CH$_4$/N$_2$ selectivity, CO$_2$/N$_2$ selectivity, and working capacity when available. For a given property, the model was trained using the mean squared error loss:
\begin{equation}
\mathcal{L}
=
\frac{1}{N}
\sum_{i=1}^{N}
\left(
\hat{y}_i - y_i
\right)^2,
\end{equation}
where $N$ is the number of training samples, $y_i$ is the reference property value, and $\hat{y}_i$ is the predicted value from the Transformer model.

During LEMO Agent optimization, the trained predictors were kept fixed and used only as fast surrogate oracles. Valid generated MOFids were evaluated by the corresponding task-specific predictor, whereas invalid candidates were assigned zero score without model inference.

\subsection{LEMO Agent optimization workflow}
\label{sec:methods_agent}

LEMO Agent performs property-guided MOF design through a closed-loop generate--validate--predict--feedback workflow. For each gas-separation task, the agent starts from a task-specific prompt that defines the target property, MOFid syntax requirements, chemical design rules, and a small set of reference MOFs. These reference MOFs are used only as examples of valid syntax and useful structural motifs, and are excluded from the final generated-candidate pool.

The optimization objective is to search the valid MOFid space for candidates with high task-specific separation performance:
\begin{equation}
x^{*}
=
\underset{x \in \mathcal{X}_{\mathrm{valid}}}{\arg\max}
\, S_t(x),
\end{equation}
where $x$ denotes a generated MOFid candidate, $\mathcal{X}_{\mathrm{valid}}$ is the valid MOFid space, $t$ denotes the target gas-separation task, and $S_t(x)$ is the task-specific optimization score.

Before iterative optimization, the agent first performs an initial generation step. The LLM is required to return a JSON object containing a fixed number of candidates. Each candidate contains a MOFid string, a short design rationale, and the generation rules used. The generated MOFids are parsed, standardized, validated, and evaluated by the corresponding Transformer-based property predictors. The evaluated initial candidates are then used to initialize the agent memory.

At optimization iteration $k$, the agent uses the current memory state $\mathcal{M}_k$ to guide the next generation step. The planner first summarizes the current optimization state and produces a design plan:
\begin{equation}
p_k
=
\operatorname{Planner}_{\phi}
\left(
t, \mathcal{M}_k
\right),
\end{equation}
where $p_k$ describes promising structural motifs, patterns to avoid, and the search direction for the next iteration. The generator then uses the design plan and memory context to propose a new batch of MOFid candidates:
\begin{equation}
\begin{aligned}
\mathcal{B}_k
&=
\operatorname{Generator}_{\phi}
\left(
t, p_k, \mathcal{M}_k
\right)  \\
&=
\left\{
x_{k,1}, x_{k,2}, \ldots, x_{k,B}
\right\},
\end{aligned}
\end{equation}
where $B$ is the number of MOFid candidates generated in each batch, and $\phi$ denotes the LLM parameters.

Each proposed MOFid is checked by the validator before property prediction. The validator enforces the required MOFid-v1 format, including the presence of a complete \texttt{MOFid-v1} suffix, a valid topology field, valid fragment ordering, and a metal-containing fragment before the suffix. Candidates with incomplete MOFid strings, missing metal fragments, unsupported topology labels, or invalid fragment ordering are treated as invalid and are not passed to the property predictors.

For each valid candidate $x$, the task-specific predictors estimate the primary selectivity and working capacity:
\begin{equation}
\begin{aligned}
\hat{\mathbf{y}}^{(t)}(x)
&=
\left(
\hat{y}_{\mathrm{sel}}^{(t)}(x),
\hat{y}_{\mathrm{wc}}^{(t)}(x)
\right)   \\
&=
\left(
f_{\mathrm{sel}}^{(t)}(x),
f_{\mathrm{wc}}^{(t)}(x)
\right),
\end{aligned}
\end{equation}
where $f_{\mathrm{sel}}^{(t)}$ and $f_{\mathrm{wc}}^{(t)}$ denote the trained predictors for selectivity and working capacity, respectively.

The optimization follows a threshold-first scoring strategy. A candidate is considered successful only when it is valid and its predicted selectivity exceeds the task-specific threshold $\tau_t$. The final optimization score is defined as
\begin{equation}
S_t(x)
=
\begin{cases}
\alpha_t
\hat{y}_{\mathrm{sel}}^{(t)}(x)
+
\beta_t
\hat{y}_{\mathrm{wc}}^{(t)}(x),
&
x \in \mathcal{X}_{\mathrm{valid}}
\ \mathrm{and}\
\hat{y}_{\mathrm{sel}}^{(t)}(x) > \tau_t,
\\
0,
&
\mathrm{otherwise},
\end{cases}
\end{equation}
where $\alpha_t$ and $\beta_t$ are task-specific weights. This scoring function prioritizes candidates that satisfy the primary selectivity requirement while further favoring candidates with higher working capacity.

After prediction and scoring, each candidate is converted into a structured feedback record containing its validity status, predicted properties, score, and feedback summary. Successful candidates are written into success memory, while invalid or low-performing candidates are written into failure memory. Recent candidates are also retained to reduce repeated generation. The memory update is written as
\begin{equation}
\mathcal{M}_{k+1}
=
\operatorname{Update}
\left(
\mathcal{M}_{k},
\mathcal{B}_{k},
\mathcal{F}_{k}
\right),
\end{equation}
where $\mathcal{F}_{k}$ denotes the structured feedback records obtained from validation, prediction, and scoring at iteration $k$.

The updated memory is then used in the next planning step, allowing the agent to reuse useful motifs, avoid repeated invalid patterns, and progressively shift the search toward high-performing regions of MOFid space. After all iterations, the final generated pool is obtained from candidates proposed by the agent during the closed-loop search. Final analyses are performed after MOFid validation, duplicate removal, and exclusion of reference examples.

\subsection{Structured memory and multi-island exploration}
\label{sec:methods_memory_islands}

A key component of LEMO Agent is the structured memory mechanism, which provides the language model with explicit feedback from previous optimization steps. Instead of treating the full generation history as an unstructured text record, we organize the search history into several memory modules:
\begin{equation}
\mathcal{M}_k
=
\left\{
\mathcal{M}_{\mathrm{ref}},
\mathcal{M}_{\mathrm{succ},k},
\mathcal{M}_{\mathrm{fail},k},
\mathcal{M}_{\mathrm{hist},k}
\right\},
\end{equation}
where $\mathcal{M}_{\mathrm{ref}}$ denotes the reference memory, $\mathcal{M}_{\mathrm{succ},k}$ denotes the success memory, $\mathcal{M}_{\mathrm{fail},k}$ denotes the failure memory, and $\mathcal{M}_{\mathrm{hist},k}$ denotes the recent-history memory at iteration $k$.

The reference memory contains high-performing MOF examples from known datasets or literature. These examples are used only as syntactic and structural anchors for the language model and are not counted as generated candidates. The success memory stores generated MOFids that pass validation and satisfy the task-specific performance criterion. The failure memory stores invalid or low-performing candidates, including malformed MOFids, candidates with unsupported topology or cat labels, and valid but low-scoring structures. The recent-history memory records candidates proposed in recent iterations to reduce duplicate generation and encourage exploration of new organic--metal--topology combinations.

After each iteration, the memory is updated according to the validation and scoring results. The success memory is refreshed with the best-performing valid candidates, the failure memory stores representative invalid or low-scoring candidates, and the recent-history memory keeps a short record of newly generated candidates. Since each memory component has a limited capacity, older or less informative records are gradually replaced by more useful feedback from later iterations.

To further reduce premature convergence, we use a multi-island exploration strategy. The full optimization process contains $K$ islands, each maintaining an independent local memory:
\begin{equation}
\mathcal{I}_j
=
\left\{
\mathcal{M}_{\mathrm{ref}},
\mathcal{M}^{(j)}_{\mathrm{succ}},
\mathcal{M}^{(j)}_{\mathrm{fail}},
\mathcal{M}^{(j)}_{\mathrm{hist}}
\right\},
\quad
j = 1,2,\ldots,K.
\end{equation}
Here, $\mathcal{M}_{\mathrm{ref}}$ is shared across islands, whereas the success, failure, and recent-history memories are maintained locally for each island. Different islands therefore accumulate different successful and failed structural patterns, allowing them to explore distinct regions of the MOFid design space.

During the initial stage, islands are selected sequentially so that each island receives at least one update and starts from its own local trajectory. After this initialization stage, island selection follows a score-based probabilistic strategy. For island $j$, the island score is defined as the best candidate score found in that island:
\begin{equation}
s_j
=
\max_{x \in \mathcal{H}_j}
S_t(x),
\end{equation}
where $\mathcal{H}_j$ denotes all candidates generated by island $j$. Islands that have discovered higher-scoring candidates are more likely to be selected in subsequent iterations. The score-based selection probability is computed using a temperature-scaled softmax:
\begin{equation}
q_j
=
\frac{
\exp\left((s_j - s_{\max})/T\right)
}{
\sum_{l=1}^{K}
\exp\left((s_l - s_{\max})/T\right)
},
\quad
s_{\max}
=
\max_{1 \leq l \leq K} s_l,
\end{equation}
where $T$ is a temperature parameter controlling the sharpness of island selection. A larger $T$ produces a smoother distribution and allows more islands to participate in the search, whereas a smaller $T$ concentrates the search on high-performing islands.

To prevent the search from being dominated too early by a single high-scoring island, we mix the score-based probability with a uniform exploration term:
\begin{equation}
p_j
=
(1 - p_{\mathrm{explore}}) q_j
+
p_{\mathrm{explore}} \frac{1}{K},
\end{equation}
where $p_{\mathrm{explore}}$ controls the strength of island-level exploration. This design assigns higher probability to islands that have discovered promising structural motifs, while still allowing lower-scoring islands to continue exploring alternative linker, metal, and topology combinations. In practice, the selected island provides its local memory to the planner and generator, and only this island is updated after the newly generated candidates are validated and scored.

In addition, LEMO Agent includes a periodic island reinitialization mechanism. After a predefined reset interval, the island with the lowest best-so-far score is reinitialized, while the best candidate from the highest-performing island is used as a founder seed for the new island. The reinitialized island retains the shared reference memory but starts a new local search trajectory from this promising candidate. This mechanism allows persistently low-performing islands to escape unproductive regions while preserving the overall multi-island exploration structure.

\subsection{Experimental Setup}\label{sec: exp_setup}
\subsubsection{Baseline methods}
\label{sec:methods_baselines}

We compared LEMO Agent with four baseline and ablated methods under the same MOFid validation and property-evaluation pipeline.

The first baseline is a direct LLM generation baseline. In this setting, the same language model was prompted with the target gas-separation task, MOFid syntax requirements, and chemical design rules, and was then asked to directly generate MOFid candidates in a single step. No iterative feedback, structured memory, or property-guided refinement was used. This baseline evaluates whether prompt-based generation alone is sufficient for producing valid and high-performing MOFs.

The second baseline is a vanilla agent. This baseline follows a basic iterative generate--evaluate--feedback loop. At each iteration, the language model generates a batch of MOFid candidates, and the same validation tools and Transformer-based property predictors used in LEMO Agent are applied to evaluate these candidates. The predicted scores and validation feedback are returned to the model to guide the next round of generation. However, unlike LEMO Agent, the vanilla agent does not use structured success/failure memory, multi-island exploration, or island-level selection. This baseline is used to distinguish the effect of simple scalar feedback from the full memory-guided optimization strategy.

The third baseline is LEMO Agent without multi-island exploration. This ablated variant keeps the same planner--generator workflow, MOFid validation, property-predictor feedback, and structured memory design as the full LEMO Agent, but uses only a single island throughout optimization. Therefore, all generated candidates are stored in one shared memory trajectory, and no island-level selection, independent local memory, or island reinitialization is used. This comparison evaluates whether maintaining multiple independent memory pools improves exploration and reduces premature convergence in MOFid space.

The fourth baseline is MOFGPT, a MOFid-based generative model for MOF design. MOFGPT is pretrained on a large-scale MOF dataset to learn the distribution of valid MOFid strings and can be further optimized by reinforcement learning to generate MOFs with desired properties. In our experiments, we used the same Transformer-based property predictor as the reinforcement-learning scoring function, so that MOFGPT was optimized toward high task-specific selectivity under the same surrogate evaluation criterion used for LEMO Agent. This setup provides a comparison with a task-adapted generative model that relies on pretraining and reinforcement-learning fine-tuning rather than language-agent planning, structured memory, and multi-island search.

\subsubsection{Evaluation metrics}
\label{sec:methods_metrics}

We evaluated both the property predictors and the generated MOF candidates. For the Transformer-based property predictors, model performance was assessed using mean absolute error (MAE) and the coefficient of determination ($R^2$). Given a test set with $N$ samples, the MAE was calculated as
\begin{equation}
\mathrm{MAE}
=
\frac{1}{N}
\sum_{i=1}^{N}
\left|
y_i-\hat{y}_i
\right|,
\end{equation}
where $y_i$ and $\hat{y}_i$ denote the reference and predicted property values of sample $i$, respectively. The $R^2$ score was calculated as
\begin{equation}
R^2
=
1-
\frac{
\sum_{i=1}^{N}
\left(y_i-\hat{y}_i\right)^2
}{
\sum_{i=1}^{N}
\left(y_i-\bar{y}\right)^2
},
\end{equation}
where $\bar{y}$ is the mean value of the reference labels in the test set.

For generation experiments, all generated candidates were first processed by the same MOFid validation pipeline. Only valid MOFids were retained for the calculation of generated-candidate statistics. For each method, we reported the average predicted selectivity of valid generated MOFs:
\begin{equation}
\overline{S}_{\mathrm{sel}}
=
\frac{1}{N_{\mathrm{valid}}}
\sum_{i=1}^{N_{\mathrm{valid}}}
\hat{y}_{\mathrm{sel}}(g_i),
\end{equation}
where $g_i$ is a valid generated MOF and $\hat{y}_{\mathrm{sel}}(g_i)$ is its predicted task-specific selectivity. When working capacity was available, we also reported the average predicted working capacity:
\begin{equation}
\overline{S}_{\mathrm{wc}}
=
\frac{1}{N_{\mathrm{valid}}}
\sum_{i=1}^{N_{\mathrm{valid}}}
\hat{y}_{\mathrm{wc}}(g_i).
\end{equation}

We further evaluated the diversity of the valid generated MOFs. For each generated set $A$, the diversity score was calculated as the average pairwise distance between valid candidates:
\begin{equation}
\mathrm{Div}(A)
=
\frac{2}{|A|(|A|-1)}
\sum_{i<j,\ g_i,g_j\in A}
D(g_i,g_j),
\end{equation}
where $D(g_i,g_j)$ denotes the distance between two valid generated MOFs. A higher diversity score indicates broader exploration of the MOFid design space.

For repeated experiments, the average selectivity, working capacity, and diversity were first calculated for each independent run and then reported as the mean and standard deviation across runs.

\subsubsection{Computational implementation}
\label{sec:methods_implementation}

All data preprocessing, MOFid validation, candidate evaluation, agent optimization, and downstream analyses were implemented in Python. RDKit was used for organic-fragment parsing, canonicalization, molecular fingerprint calculation, and cheminformatics-based analyses. The Transformer-based property predictors were implemented in PyTorch and used as fixed surrogate models during the LEMO Agent optimization process.

The language model was accessed through the OpenAI API using GPT-5.1. For both initialization and iterative optimization, the model was required to return JSON-formatted outputs. All experiments were conducted on an Ubuntu Linux system equipped with eight NVIDIA GeForce RTX 4090 GPUs.

\section*{Competing Interests}
The authors declare no competing interests.

\bibliography{reference}

\newpage
\begin{appendix}

\captionsetup[figure]{labelformat=empty}
\captionsetup[table]{labelformat=empty}
\renewcommand{\thefigure}{Supplementary Figure \arabic{figure}}
\renewcommand{\thetable}{Supplementary Table \arabic{table}}
\setcounter{figure}{0}
\setcounter{table}{0}

\end{appendix}
\end{document}